%% file: arxiv.tex
\documentclass[10pt,twocolumn,letterpaper]{article}

\usepackage[pagenumbers]{cvpr} 


\usepackage{algorithm}
\usepackage{algorithmic}

\definecolor{cvprblue}{rgb}{0.21,0.49,0.74}
\usepackage[pagebackref,breaklinks,colorlinks,allcolors=cvprblue]{hyperref}

\usepackage{multirow}
\usepackage{subcaption}

\usepackage{colortbl}
\usepackage{xcolor}

\title{PBDyG: Position Based Dynamic Gaussians\\for Motion-Aware Clothed Human Avatars}

\author{Shota Sasaki$^{1}$\quad Jane Wu$^{2}$ \quad Ko Nishino$^{1}$\\
$^{1}$ Kyoto University\qquad  $^{2}$ University of California, Berkeley \\
\texttt{\small \href{https://vision.ist.i.kyoto-u.ac.jp/research/pbdyg/}{https://vision.ist.i.kyoto-u.ac.jp/research/pbdyg/}}
}

\begin{document}

\twocolumn[{%
\renewcommand\twocolumn[1][]{#1}%
\maketitle
\begin{center}
    \centering
    \captionsetup{type=figure}
    \includegraphics[width=\linewidth]{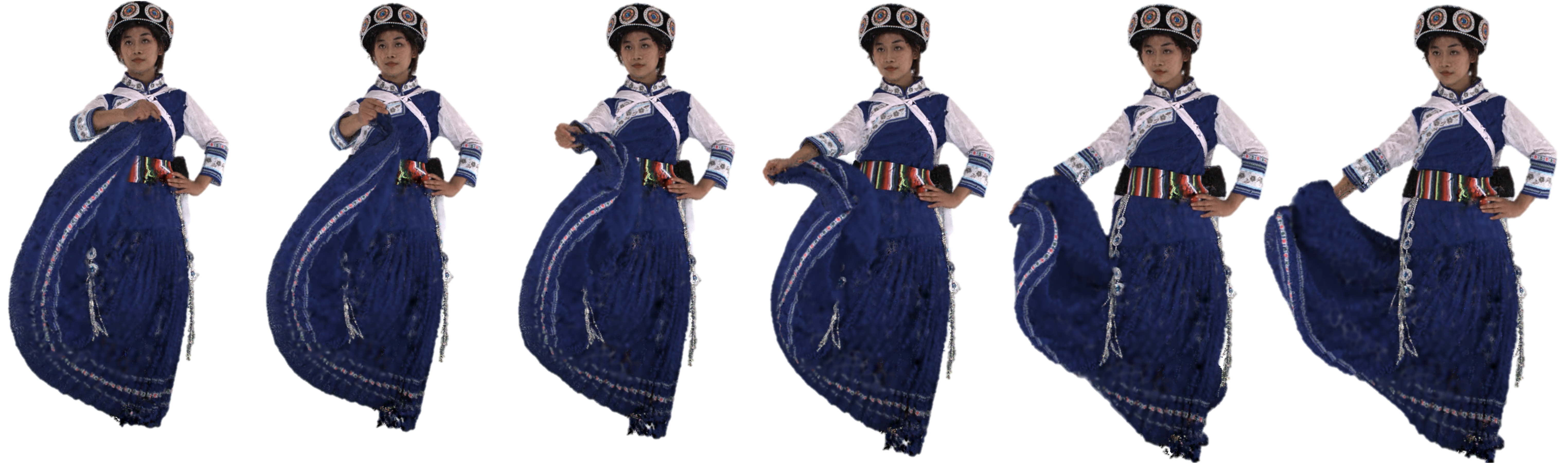}
    \captionof{figure}{Position Based Dynamic Gaussians (PBDyG) reconstructs a person from multiview videos such that their clothing deforms in accordance with the \textit{movement}, not just rigid pose, of the person. This makes possible reconstruction of animatable avatars of people wearing highly deformable loose clothes. Example test animations are shown.}
\end{center}%
}]

\begin{abstract}
This paper introduces a novel clothed human model that can be learned from multiview RGB videos, with a particular emphasis on recovering physically accurate body and cloth movements. Our method, Position Based Dynamic Gaussians (PBDyG), realizes ``movement-dependent'' cloth deformation via physical simulation, rather than merely relying on ``pose-dependent'' rigid transformations. We model the clothed human holistically but with two distinct physical entities in contact: clothing modeled as 3D Gaussians, which are attached to a skinned SMPL body that follows the movement of the person in the input videos. The articulation of the SMPL body also drives physically-based simulation of the clothes' Gaussians to transform the avatar to novel poses. In order to run position based dynamics simulation, physical properties including mass and material stiffness are estimated from the RGB videos through Dynamic 3D Gaussian Splatting. Experiments demonstrate that our method not only accurately reproduces appearance but also enables the reconstruction of avatars wearing highly deformable garments, such as skirts or coats, which have been challenging to reconstruct using existing methods.
\end{abstract}

\section{Introduction}
Digitization of real-world human behavior, \ie, the 3D modeling of people and her movements that allows for post-capture interaction, is a fundamental building block of many visual applications. Its realism is essential for virtual communication in VR/AR and for sensing and planning in robotics. Accurate 4D modeling of human behavior is also critical for gauging the subtle internals of the person such as her condition and mood from passive observations, which underlies societal applications including elderly care, security, and commerce, to name a few. 

The computer vision community has made large strides in human behavior modeling research, starting with joint detection~\cite{sarafianos20163d}. The invention of heat maps led to robust 2D skeletal pose estimation. Introduction of statistical priors on the human body (\eg, SMPL~\cite{loper2015smpl}) has enabled posed body surface reconstruction. With the advent of volume-rendering based appearance representations, \eg, NeRF~\cite{mildenhall2021nerf} and 3D Gaussian Splatting (3DGS)~\cite{qian20243dgs}, recent works have realized image-based clothed human body reconstruction~\cite{liu2021neural,peng2021neural,peng2021animatable,kwon2021neural,suo2021neuralhumanfvv,zhao2022humannerf,kocabas2024hugs,qian20243dgs,xu2024relightable,li2024animatable,moreau2024human}.
These human ``avatars'' can then be reposed via rigid skinning of the cloth and body.

Two fundamental limitations, however, hamper existing methods. One is that they are ``pose-dependent'' but not ``movement dependent,'' such that cloth dynamics are modeled in a physically-based manner.  Imagine a person wearing a long skirt. As the person swings her body, the skirt will follow but its shape would depend on momentum, \ie, the direction the body is heading, even for the exact same skeletal pose. Current methods are unable to model this movement-dependent garment behavior because the shape-outputting network is typically only conditioned on the skeletal pose of the body.

The second limitation is that existing approaches model clothing and body as separated physical quantities. This separation has its advantage, as it is convenient for composition, \eg, redressing a person. Gaussian Garments~\cite{rong2024gaussian} is one such method that models clothes as a pre-acquired template deformed with a graph neural network. This disentangled modeling of the body and clothing, however, makes representing the interaction between them challenging. In the real world, clothing interacts with the body via complex friction and collision forces that are highly movement dependent.
These forces cannot be accurately modeled with basic contact forces at a few points, making it very challenging to simulate let alone extract from visual observations when the body and its clothes are separated.

How then can we model clothed human behavior complete with movement-dependent clothes motion driven by accurate body and clothes interactions? In other words, how can we extract a truly animatable 4D human model from passive observations---a holistic clothed human model that can be gesticulated in arbitrary body movements? 

We achieve this by learning both the body and the clothes as a holistic animatable object directly from a multiview video. Unlike past methods, we do not require pre-scanned templates of the clothing~\cite{li2024diffavatar}, nor do we utilize ground truth meshes of the body and clothing~\cite{zheng2024physavatar}.
Our key idea is to model the clothed body as a whole, but also as two distinct physical entities in contact. We model deformable clothing with 3DGS associated with the surface of SMPL which models the body. In other words, the Gaussian clothing is rigged with a SMPL body whose movements drive the physically-based simulation of the Gaussians of the clothes. We employ position based dynamics (PBD)~\cite{muller2007position} for simulating the clothes' movements. The physical properties including mass and material stiffness are estimated for each of the clothes surface points (\ie, Gaussians). 
That is, we derive and dress the SMPL human model with these \textbf{P}osition \textbf{B}ased \textbf{Dy}namic \textbf{G}aussians (PBDyG) and learn their parameters from a multiview video.

Given synchronized RGB videos capturing a person from multiple views, learning PBDyG consists of 3 main steps. First, we select a frame close to a T-pose and perform 3D reconstruction using 3DGS for this frame. Next, we track the set of Gaussians obtained from the 3D reconstruction using the tracking method proposed in Dynamic 3D Gaussians~\cite{luiten2023dynamic}. Through this process, we acquire the trajectory of each 3D Gaussian composing the person across all frames.
Using the Gaussian trajectories obtained thus far and the refined SMPL annotations, we propose a novel method to construct an animatable human avatar that can be driven by a combination of linear blend skinning (LBS) and PBD. Furthermore, we optimize the physical parameters of PBD to ensure that the avatar accurately reproduces the tracking results captured by Dynamic 3D Gaussians.

Experiments demonstrate that our method not only reconstructs geometry and appearance that are faithful to the input videos but also enables the reconstruction of avatars wearing highly deformable garments, such as loose fitting skirts or coats.
Because our method estimates motion-dependent physical parameters and disentangles the differing properties of the human body and clothing while preserving complex contact, we are able to generate simulation-ready 3D geometry and appearance for reanimation and integration with downstream simulation engines. We plan to release all code and data to encourage the use of PBDyG as a foundational tool for such applications.

\section{Related Work}
\noindent\textbf{Neural Rendering for Human Reconstruction.}
Neural rendering 3D representations such as Neural Radiance Fields (NeRFs) \cite{mildenhall2021nerf} and 3D Gaussian Splatting (3DGS) \cite{kerbl20233d} have recently become popular for creating and animating digital avatars from multiview video.
Following the release of NeRF, the earliest works in this area \cite{liu2021neural,peng2021neural,peng2021animatable,kwon2021neural,suo2021neuralhumanfvv,zhao2022humannerf} rely on the SMPL model \cite{loper2015smpl} to skin an optimized NeRF of a human from a canonical space (unskinned) space to the observation space (skinned) and vice versa, which is key to both animation and dynamic reconstruction.
HumanNeRF \cite{weng2022humannerf} optimizes a learned motion field in conjunction with a NeRF of the person in a canonical pose.
This network-inferred skinning thus overcomes clothing artifacts introduced by directly using SMPL.
TAVA \cite{li2022tava} additionally adds learned shading effects conditioned on skeletal pose to parameterize appearance.
ARAH \cite{wang2022arah} introduces a novel joint root-finding algorithm for simultaneous ray-surface
intersection search and correspondence search, which improves the reanimation quality of the optimized NeRF.
MonoHuman \cite{yu2023monohuman} focuses on the monocular video setting and proposes the use of a shared bidirectional deformation module to impose cycle consistency and a correspondence search module to enforce consistency between frames.
HumanRF \cite{icsik2023humanrf} aims to close the gap in production-level quality by synthesizing humans at 12MP and introduces the ActorsHQ dataset captured using 160 cameras per example sequence.

\noindent\textbf{Pose-Dependent Human Avatars.}
More recently, a number of 3DGS-based approaches have been proposed for creating animatable avatars from monocular \cite{hu2024gaussianavatar,hu2024gauhuman,lei2024gart} and multiview \cite{kocabas2024hugs,qian20243dgs,xu2024relightable,li2024animatable,moreau2024human} video.
GaussianAvatar \cite{hu2024gaussianavatar} uses pixel-aligned features for optimizing 3D Gaussians from monocular video.
Animatable Gaussians \cite{li2024animatable} takes multiview videos as input and leverages the structure of CNNs by constructing front and back canonical Gaussian maps.

In addition to 3D reconstruction, a number of recent works aim to recover physical properties of the human body and clothing from videos.
Dressing Avatars \cite{xiang2022dressing} proposes a clothing appearance model that operates on top of explicit geometry, thereby preserving appearance while being grounded in high quality geometry.
PhysAvatar \cite{zheng2024physavatar} also relies on knowing the ground truth human and cloth geometry, and the proposed method builds on a physics-based simulator and rendering engine to optimize 4D Gaussians.
IntrinsicAvatar \cite{wang2024intrinsicavatar} uses explicit Monte Carlo ray tracing to model secondary shadow effects.
To our knowledge, our work is the first to propose learning physical properties of clothed human avatars modeled using 3DGS.

\noindent\textbf{Disentangling Body and Cloth.}
Recovering the individual garments worn on the body is important not only for performance capture, but also for animation and re-targeting.
DiffAvatar \cite{li2024diffavatar} generates simulation-ready cloth from multiview video by optimizing a template garment's design and material parameters, along with SMPL body shape.
Gaussian Garments \cite{rong2024gaussian} begins with reconstructing the garment from a single multiview frame, and then optimizing geometry and appearance across time.
Most similar to our work, AniDress \cite{chen2024anidress} proposes a garmenet rigging model to handle pose-dependent cloth deformation separately from the rigid transformations of the body. Our method, PBDyG, disentangles the body and clothes as distinct physical entities but preserves their contacts as the association between them, which is achieved by ``rigging'' the 3DGS clothing with the SMPL body. This disentangled but holistic modeling is essential for modeling the body motion-dependent cloth movements.

\begin{figure*}[ht]
    \centering
    \includegraphics[width=\linewidth]{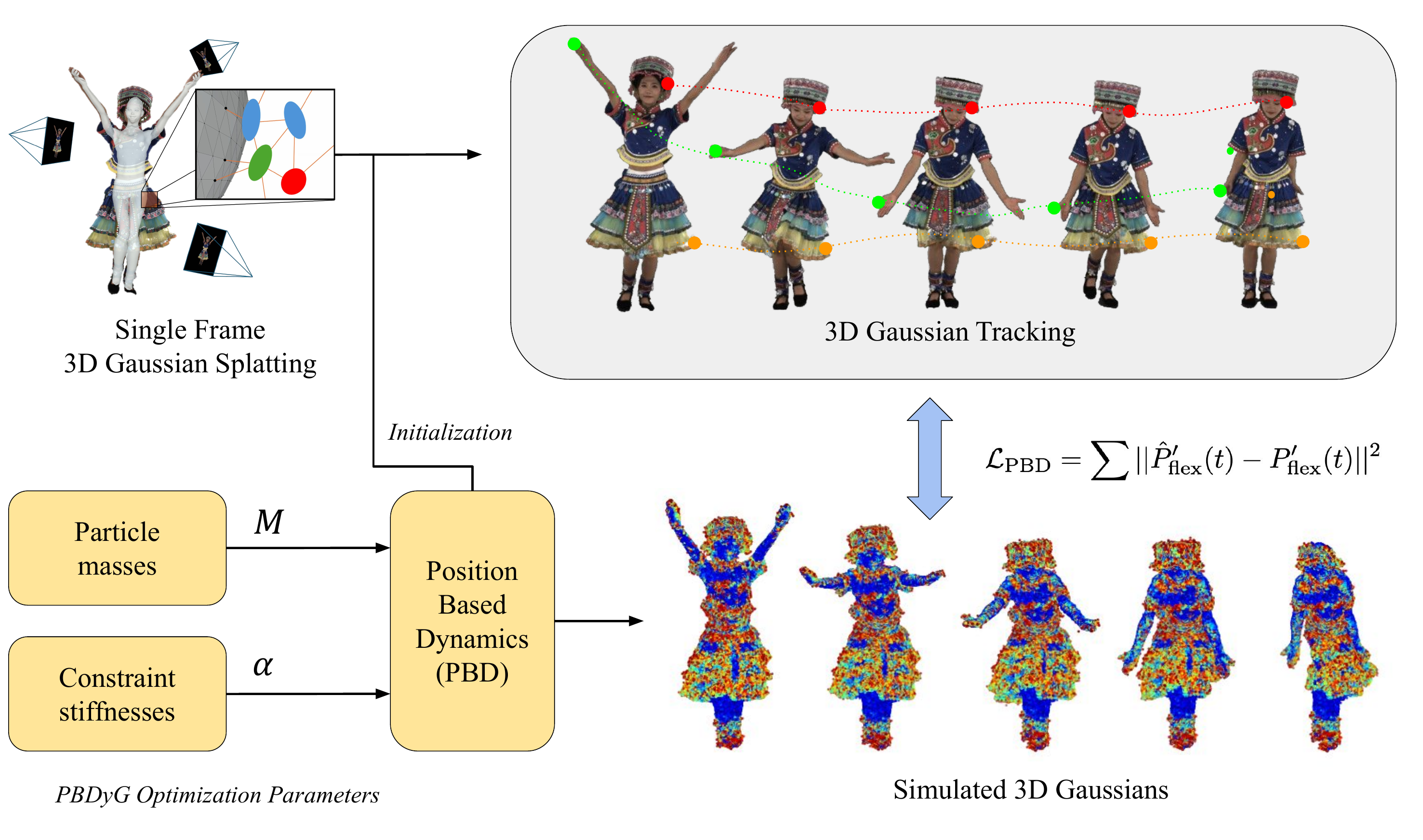}
    \caption{Overview of our method. PBDyG learning consists of three main steps:  3D Gaussian reconstruction and tracking, SMPL fitting and refinement also referencing the Gaussians, and PBD parameter estimation.}
    \label{fig:enter-label}
\end{figure*}

\section{Position Based Dynamic Gaussians}
Given multiview RGB videos of a human in motion, we aim to recover not only the geometry and appearance of the performance, but also physical properties of both the body and clothing for reanimation. 
We aim to model the movement-dependent behavior of clothing; this involves not only predicting pose-dependent rigid transformation, but also dynamics-based deformation.
Many existing 3DGS-based avatar reconstruction methods use linear blend skinning (LBS)~\cite{lander1998skin,magnenat1988joint} to change the pose of an avatar~\cite{kocabas2024hugs,qian20243dgs,xu2024relightable}. However, intense deformations, such as those occurring in skirts or coats, cannot be adequately represented using a rigid deformation model.
To address this limitation, our method introduces a physical simulator (PBD) to model such deformations. By incorporating PBD, our approach effectively captures the motion-dependent, dynamic deformations of clothing.

Our method consists of three main steps: reconstructing and tracking 3D Gaussians (Section \ref{sec:dynamic_gaussians}, fitting and refining the SMPL model to the video while also referencing the 3D Gaussians (Section \ref{sec:smpl_fitting}), and learning PBD body and cloth parameters to be used for reanimation  (Section \ref{sec:pbdyg}).

\subsection{Dynamic Human Gaussians}\label{sec:dynamic_gaussians}
Given $T$ frames of RGB multiview images $I_{1:T, 1:C}$ captured from $C$ viewpoints, along with estimated SMPL parameters at each time step, $\theta_{1:T}$ and $\beta_{1:T}$, our first goal is to obtain the trajectories of a set of Gaussians, $G_{1:T}$, which will later serve as references for SMPL refinement and PBD parameter optimization.

\noindent\textbf{3D Gaussian Splatting.}
In 3D Gaussian Splatting \cite{kerbl20233d}, the scene is represented by a collection of 3D Gaussians, each of which is parameterized by its position $\mu$, rotation $q$, scale $s$, color $c$ and opacity $\sigma$. Each 3D Gaussian is defined as 
\begin{equation}
G(x)=
\sigma
\exp(- \frac{1}{2}(x - \mu)^T\Sigma^{-1}(x - \mu))\,,
\end{equation}
where $\Sigma = RSS^TR^T$ is the anisotropic covariance matrix that controls the shape of the Gaussian. For rendering, the covariance matrix is projected onto image space with the camera view transform. The 2D projection of the covariance matrix $\Sigma'$ becomes
$\Sigma^{'} = J W \Sigma W^T J^T$, where $W$ is the view transform and $J$ is the Jacobian of the affine approximation of the projective transform. The resulting 2D splat is used for alpha blending, allowing multiple overlapping Gaussians to contribute to the final pixel color.

Using $\mathcal{I}_{1,1:C}$, we optimize the parameters of 3DGS to obtain a set of Gaussians $G$ that models the clothed person captured in the video.

\noindent\textbf{Dynamic Gaussians.}
Dynamic 3D Gaussians~\cite{luiten2023dynamic} enables dynamic scene reconstruction and 6-DOF tracking. The method achieves this by representing a dynamic scene with a set of evolving 3D Gaussians that move over time to model the scene's physical and visual properties.

Dynamic 3D Gaussians employs three physically-motivated prior losses to ensure the Gaussians evolve coherently over time. The first is the local rigidity loss which enforces nearby Gaussians to move together as if they were part of a rigid body. 
The second is the rotation similarity loss which ensures that neighboring Gaussians maintain similar rotations over time.
Finally, it leverages an isometry loss which preserves the relative distances between neighboring Gaussians across timesteps so that they do not drift apart.
We use Dynamic Gaussians to track the set of Gaussians $G$ that model the avatar clothese across all frames, which provides us with the positions $G(t)$ of the Gaussians at each time step.

\subsection{SMPL Refinement}\label{sec:smpl_fitting}
Many avatar reconstruction methods rely on the SMPL model \cite{loper2015smpl} to deform the avatar, where the accuracy of the SMPL fit directly impacts the accuracy of the reconstruction. In our method, the accuracy of the SMPL model is even more critical as the clothing simulation is driven by it. Unfortunately, joint estimation with loose garments, such as skirts, is challenging as the keypoints are deeply veiled by the clothes. This easily leads to inaccurate SMPL estimates.

We overcome this dilemma with a novel SMPL refinement scheme. We use key points only from the face, where relatively accurate keypoint detection can be achieved. We define the keypoint loss $L_{\mathrm{keypoints}}$ as 
\begin{equation}
L_{\mathrm{keypoints}} = \sum_{i \in \text{face}} \| \mathbf{x}_i - \mathbf{x}_i^{gt} \|^2\,,
\end{equation}
where $\mathbf{x}_i$ and $\mathbf{x}_i^{gt}$ denote the predicted and ground truth keypoints for each facial feature, respectively.

To robustly optimize the SMPL model even in the presence of inaccurate key points, we leverage the reconstruction from Gaussian Splatting. Gaussians associated with the body are expected to be located within the mesh interior. Therefore, for each Gaussian $G_i$ belonging to the body, the nearest point $V_{\mathcal{N}(i)}$ on the SMPL mesh is calculated. To encourage alignment between the SMPL model and the results of Gaussian Splatting, we introduce an alignment loss 
\begin{equation}
\mathcal{L}_{\mathrm{align}} = \sum_{i \in \text{body}} \| G_i - V_{\mathcal{N}(i)} \|^2\,,
\end{equation}
which promotes alignment of SMPL with the Gaussian Splatting results by minimizing the distance between each body-associated Gaussian and its nearest mesh point.

These two losses alone can lead to trivial solutions where the SMPL model expands excessively to enclose the entire set of Gaussians. To avoid this, we introduce an additional loss that encourages each projected vertex of the SMPL model to remain inside the human mask $\mathcal{M}$
\begin{equation}
\mathcal{L}_{\mathrm{verts}} = \sum_j (1 - \mathcal{M}(\Pi(V_j))\,,
\end{equation}
where $\Pi$ denotes the camera projection for each vertex $V_j$

To further prevent extreme solutions, we impose regularization on the SMPL pose and shape parameters
\begin{equation}
\mathcal{L}_{\mathrm{preg}} = \sum \|\theta\|^2\,,
\mathcal{L}_{\mathrm{sreg}} = \sum \|\beta\|^2\,.
\end{equation}

The complete loss $\mathcal{L}_{\mathrm{init}}$ for optimizing the SMPL parameters in the initial frame is
\begin{equation}
\mathcal{L}_{\mathrm{init}} = \lambda_{\mathrm{align}} \mathcal{L}_{\mathrm{align}} + \lambda_{\mathrm{verts}} \mathcal{L}_{\mathrm{verts}} + \lambda_{\mathrm{preg}} \mathcal{L}_{\mathrm{preg}} + \lambda_{\mathrm{sreg}} \mathcal{L}_{\mathrm{sreg}}\,.
\end{equation}

For the remaining frames, the Gaussians from the initial frame are deformed using Linear Blend Skinning (LBS), and a loss function is introduced to encourage alignment between the body Gaussians and the results of Dynamic Gaussian tracking
\begin{equation}
\mathcal{L}_{\mathrm{track}} = \sum_{i \in \text{body}} \| \mathrm{LBS}(G_i(1), \theta(t), \beta) - G_i(t) \|^2\,.
\end{equation}
From the second frame onward, the shape parameter $\beta$ is held constant at its initial frame value, and only the pose parameters are optimized.

The overall loss function for optimization is 
\begin{equation}
\mathcal{L}_{\mathrm{remain}} = \lambda_{\mathrm{track}} \mathcal{L}_{\mathrm{track}} + \lambda_{\mathrm{keypoint}} \mathcal{L}_{\mathrm{keypoint}}\,.
\end{equation}
By minimizing this loss function at each time step, the SMPL parameters are optimized to align with the Gaussian tracking results over time.

\subsection{PBDyG Parameter Optimization}\label{sec:pbdyg}

\noindent\textbf{Position Based Dynamics.}
PBD~\cite{muller2007position} is a widely adopted physically-based method for real-time dynamics simulation. The core idea of PBD is to bypass explicit velocity and force computations by directly manipulating object positions to satisfy given constraints. PBD consists of three basic steps: predicting positions using current velocity; iteratively projecting the positions to meet the constraints; and updating the velocities. Please see the appendix for more details. 
PBD handles constraints efficiently through projection, making it stable for real-time applications like cloth or soft body simulations. Its main drawback, however, is the dependency of constraint stiffness on the timestep and iteration count, leading to unpredictable behavior in more complex scenarios.

Extended Position Based Dynamics (XPBD) \cite{macklin2016xpbd} was introduced to address the stiffness limitations of traditional PBD by decoupling the constraint stiffness from the number of iterations and timestep. This extension allows for consistent stiffness behavior regardless of simulation parameters, making it more physically accurate while retaining the efficiency of PBD. XPBD introduces an additional variable, the Lagrange multiplier $\lambda$, to handle compliance (\ie, flexibility in constraint satisfaction). This leads to a more robust solution of constraints by using a revised position update 
\begin{equation}
\Delta \mathbf{x} = M^{-1} \nabla C(\mathbf{x_i})^T \Delta \lambda\,,
\end{equation}
where $M$ is the mass matrix, and $\nabla C(\mathbf{x_i})$ is the gradient of the constraint function with Lagrange multiplier update
\begin{equation}
\Delta \lambda_j = 
\frac{- C_j(\mathbf{x_i}) - \tilde{\alpha}_j \lambda_{ij}}{\nabla C_j M^{-1} \nabla C^T_j + \tilde{\alpha}_j}\,,
\end{equation}
where $\alpha$ is a compliance term that controls the stiffness of the constraints. Higher $\alpha$ values lead to softer constraints, while $\alpha = 0$ corresponds to a rigid constraint. By incorporating this multiplier, XPBD maintains the benefits of PBD (efficiency and simplicity) while offering improved control over constraint stiffness, even with varying time steps and iteration counts. It is especially useful for simulating deformable materials and handling interactions between soft and rigid objects.

\noindent\textbf{Integration with Gaussian Splatting.}
When a person moves, both rigid deformations, such as changes in body pose, and non-rigid deformations, like clothing swings, occur. Our approach represents rigid deformations using Linear Blend Skinning (LBS) and non-rigid deformations using Position Based Dynamics (PBD).

To model deformations resulting from the combination of LBS and PBD, we define a point set $P$ consisting of the SMPL vertices $V$ and the reconstructed Gaussian set $G$, $P = V + G$.

Gaussian Splatting typically generates hundreds of thousands of Gaussians. Using all of these would be computationally intractable. We sample a subset of points from $P$, denoted $P^{'}$. Deformations are applied only to the sampled points $P^{'}$, and the positions of the remaining points are determined through interpolation.

To establish connectivity, we apply Delaunay Triangulation to $P^{'}$, forming a tetrahedral mesh $T$ and introducing distance constraints on each edge $E$ of the resulting tetrahedra. The sampled point set $P^{'}$ contains points derived from SMPL, $P^{'}_{\mathrm{smpl}}$, and from the Gaussians, $P^{'}_{\mathrm{gaussian}}$.  Additionally, within $P^{'}_{\mathrm{gaussian}}$, there are two types of points: those from Gaussians representing clothing, $P^{'}_{\mathrm{cloth}}$, and those from Gaussians representing the body, $P^{'}_{\mathrm{body}}$
\begin{equation}
P' = P'_{\mathrm{smpl}} + P'_{\mathrm{gaussian}}\,,
P'_{\mathrm{gaussian}} = P'_{\mathrm{body}} + P'_{\mathrm{cloth}}\,.
\end{equation}

Since it is desirable to model $P^{'}_{\mathrm{smpl}}$ and $P^{'}_{\mathrm{body}}$ with rigid deformations, these points are deformed using LBS and remain fixed during PBD simulation. In contrast, modeling $P^{'}_{\mathrm{cloth}}$ with non-rigid deformations is preferred, so the movement of these points is computed through PBD.
Here, we redefine $P'_{\mathrm{rigid}} = P'_{\mathrm{smpl}} + P'_{\mathrm{body}}$, $P'_{\mathrm{flex}} = P'_{\mathrm{cloth}}$.

\vspace{3pt}
\noindent\textbf{PBDyG Optimization.}
Given $P'(1;T)$ and the SMPL annotations $\theta(1;T)$ and $\beta$ for each frame, our objective is to optimize the parameters of PBD to accurately reproduce $P'$.
The parameters optimized in this method include the mass $M$ of each point in $P'$, the compliance $\alpha$ of each connection $E$.

For optimization, we first randomly select a time $t$, and calculate the next time step’s $P'_{\mathrm{rigid}}$ 
\begin{equation}
    \hat{P}'_{\mathrm{rigid}}(t+1) = \mathrm{LBS}(P'_{\mathrm{rigid}}(1), \theta_{t}, \beta)\,.
\end{equation}
Next, at time $t$, using the point positions $P'(t)$, velocities $\dot{P}'(t)$, the previously computed $\hat{P}'_{\mathrm{rigid}}(t+1)$, and current parameters $\alpha, M$, we compute the next time step's $P'_{\mathrm{flex}}$ through Position Based Dynamics
\begin{equation}
\hat{P}'_{\mathrm{flex}}(t+1) = \mathrm{PBD}(P'(t), \dot{P}'(t), \hat{P}'_{\mathrm{rigid}}(t+1), \alpha, M)\,.
\end{equation}

The mean squared error between the predicted position at time $t+1$ and the tracking result is given by
\begin{equation}
\mathcal{L}_{\mathrm{PBD}} = \sum || \hat{P}'_{\mathrm{flex}}(t) - P'_{\mathrm{flex}}(t) ||^2\,.
\end{equation}
By minimizing $\mathcal{L}_{\mathrm{PBD}}$, we optimize $\alpha, M$ to create a human avatar that reflects the actual material properties of the clothing rigged and driven by its SMPL body.

\noindent\textbf{Substep Strategy.}
In PBD, the position is updated by considering the current velocity and the effects of gravity, and then an optimization loop is executed to satisfy the constraints. In our setup, if the time step is too large, the gravitational fall becomes significant, making it difficult to correct within the optimization loop. To address this, we divide the frame into 10 small steps and perform forward simulation of PBD at each substep. The vertex positions of the SMPL model in each substep are obtained by linearly interpolating between frames.

\noindent\textbf{Constraints.}
We empirically discovered that with distance constraints alone, the set of Gaussians tends to collapse during the optimization of PBD parameters. To maintain the flexibility of clothing while preventing the collapse of both appearance and geometry, we use the AirMesh constraint~\cite{muller2015airmesh} in addition to the distance constraint. The AirMesh constraint is a unilateral constraint that acts only when a tetrahedron, generated by Delaunay triangulation, undergoes a flip, in order to resolve the inversion. Consequently, it helps prevent simulation instability while maintaining the flexibility of the fabric. We want to emphasize that the parameters associated with the AirMesh constraint are not part of the optimization target.

\begin{table}[t]
    \scalebox{0.65}{
    \begin{tabular}{c|lccccc}
        \toprule
        Seq & Method & PSNR $\uparrow$ & SSIM $\uparrow$ & LPIPS $\downarrow$ & HF-SSIM $\uparrow$ & HF-PSNR $\uparrow$ \\ 
        \hline
        \multirow{2}{*}{0166} & AG\cite{li2024animatable} & \cellcolor{orange!25} 20.15 & \cellcolor{orange!25} 0.932 & \cellcolor{orange!25} 0.108 & 0.897 &  14.11\\
        & \textbf{PBDyG(Ours)} & 18.36 & 0.921 & 0.121 & \cellcolor{orange!25} 0.900 & \cellcolor{orange!25} 14.13\\
         
        \midrule
        \multirow{2}{*}{0206} & AG\cite{li2024animatable} & 17.34 & \cellcolor{orange!25} 0.914 & \cellcolor{orange!25} 0.160 & 0.867 & 13.06\\
        & \textbf{PBDyG(Ours)} & \cellcolor{orange!25} 17.46 & 0.909 & 0.173 & \cellcolor{orange!25} 0.875 & \cellcolor{orange!25} 13.19 \\
         
        \midrule
        \multirow{2}{*}{0007} & AG\cite{li2024animatable} & \cellcolor{orange!25} 19.27 & \cellcolor{orange!25} 0.927 & \cellcolor{orange!25} 0.135 & 0.875 & 13.19\\
        & \textbf{PBDyG(Ours)} & 18.29 & 0.919 & 0.154 & \cellcolor{orange!25} 0.881 & \cellcolor{orange!25} 13.23\\
        \bottomrule
    \end{tabular}
    }
    \centering 
    \caption{Quantatative Comparison with SoTA Avatar reconstruction method (Animatable Gaussians \cite{li2024animatable}). Please see text for details of metrics. HF-SSIM and HF-PSNR are newly introduced to better capture the true ``accuracy" of reconstructions. Please see detailed comparisons with qualitative results in following figures.}
    \label{tab:comparison_results}
\end{table}

\captionsetup[subfigure]{labelformat=empty}

\begin{figure}[t]
    
    \begin{subfigure}[h]{0.15\textwidth}
    \centering
        \includegraphics[width=0.8\textwidth]{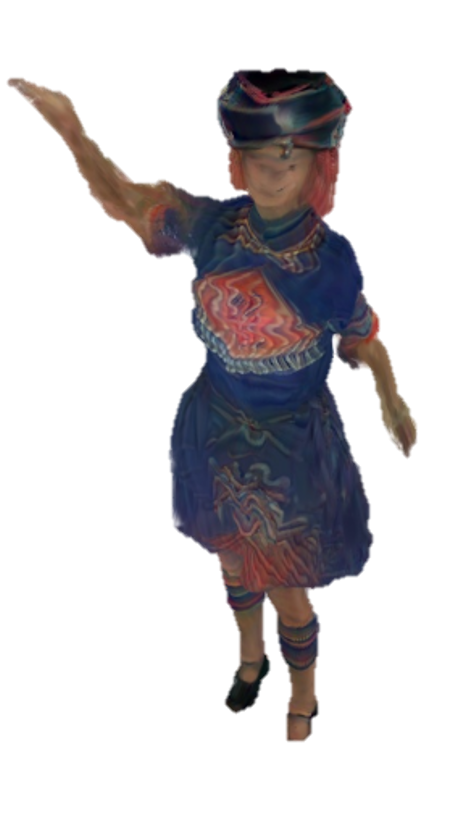}
        \subcaption{\hspace{1.1em}\mbox{Animatable Gaussians}} 
    \end{subfigure}
    \hfill
    \begin{subfigure}[h]{0.15\textwidth}
        \centering
        \includegraphics[width=0.8\textwidth]{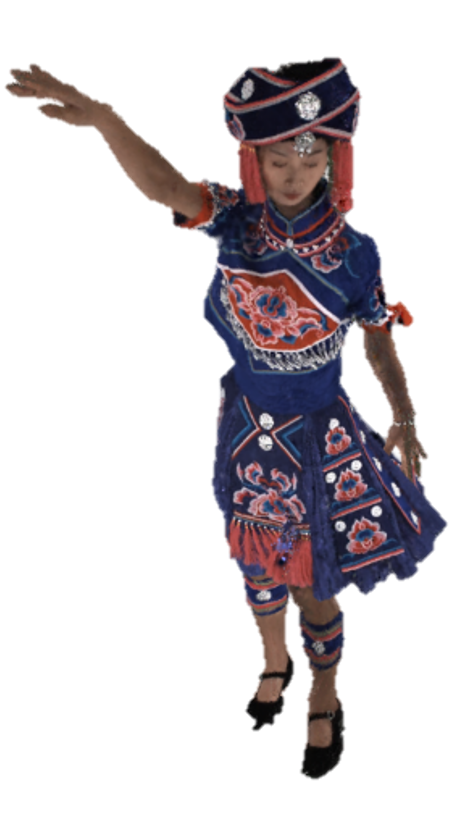}
        \caption{\hspace{1.1em}Ours} 
    \end{subfigure}
    \hfill
    \begin{subfigure}[h]{0.15\textwidth}
        \centering
        \includegraphics[width=0.8\textwidth]{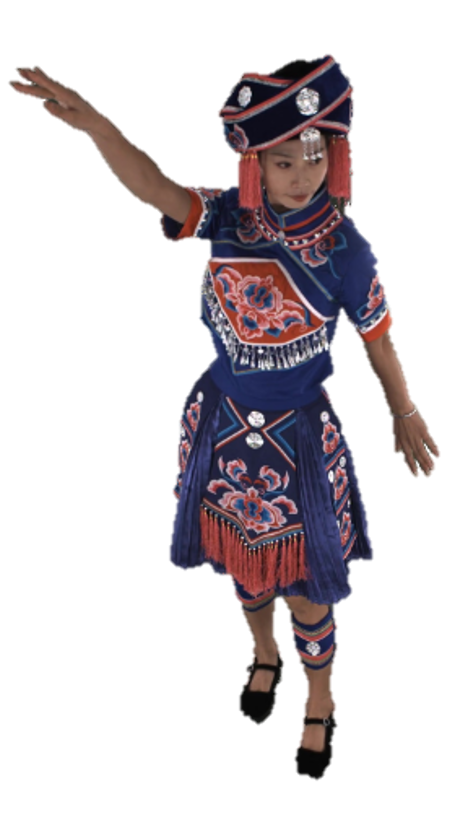}
        \caption{Ground Truth} 
    \end{subfigure}
    \par
    \vspace{10pt}
    \begin{subfigure}[h]{\linewidth}
        \scalebox{0.7}{
        \begin{tabular}{c|ccccc}
            \toprule
            Method & PSNR $\uparrow$ & SSIM $\uparrow$ & LPIPS $\downarrow$ & HF-SSIM $\uparrow$ & HF-PSNR $\uparrow$\\
            \midrule
            AG\cite{li2024animatable} & \cellcolor{orange!25} 20.44 & \cellcolor{orange!25} 0.931 & 0.120 & 0.906 & 14.61\\
            \textbf{PBDyG(Ours)} & 19.34 & 0.928 & \cellcolor{orange!25} 0.112 & \cellcolor{orange!25} 0.910 & \cellcolor{orange!25} 14.89\\
            
            \bottomrule
        \end{tabular}
        }
    \end{subfigure}
    \caption{Reconstruction comparison for Subject 0166. Our method achieves more accurate visual results yet score poorly on conventional metrics which do not reflect perceptual quality well. We believe our new metric is more faithful to the qualitative results.}
    \label{fig:0166-hf-ssim}
\end{figure}

\begin{figure}[t]
    \begin{subfigure}[h]{0.155\textwidth}
        \centering
        \includegraphics[width=0.8\textwidth]{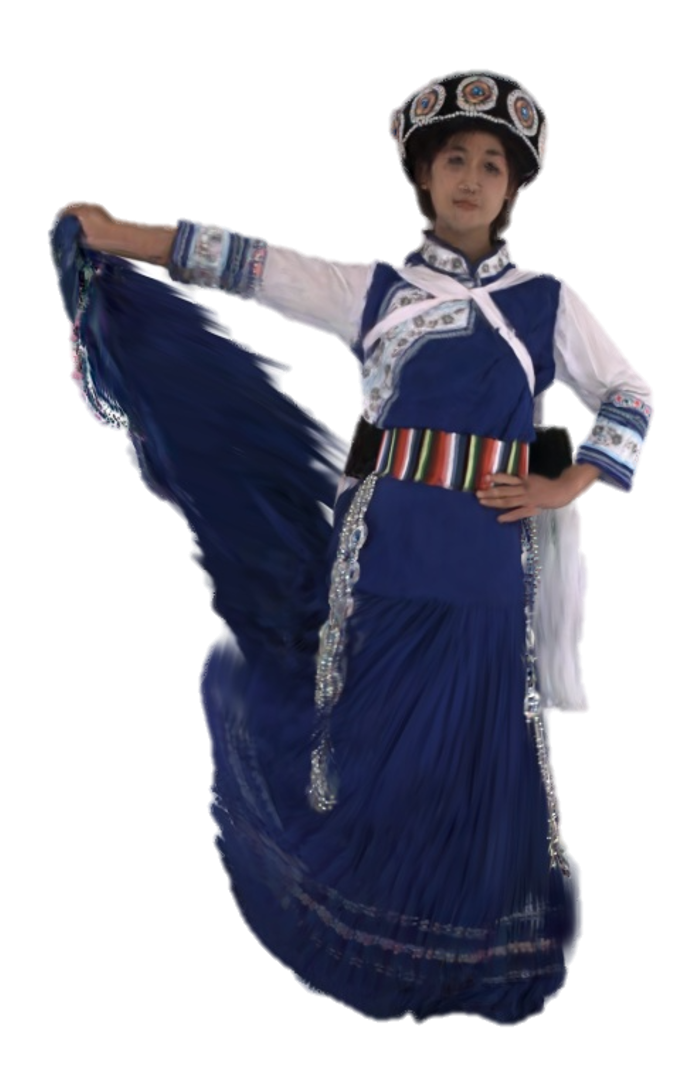}
        \caption[r]{\hspace{1.1em}\mbox{Animatable Gaussians}}
    \end{subfigure}
    \hfill
    \begin{subfigure}[h]{0.155\textwidth}
        \centering
        \includegraphics[width=0.8\textwidth]{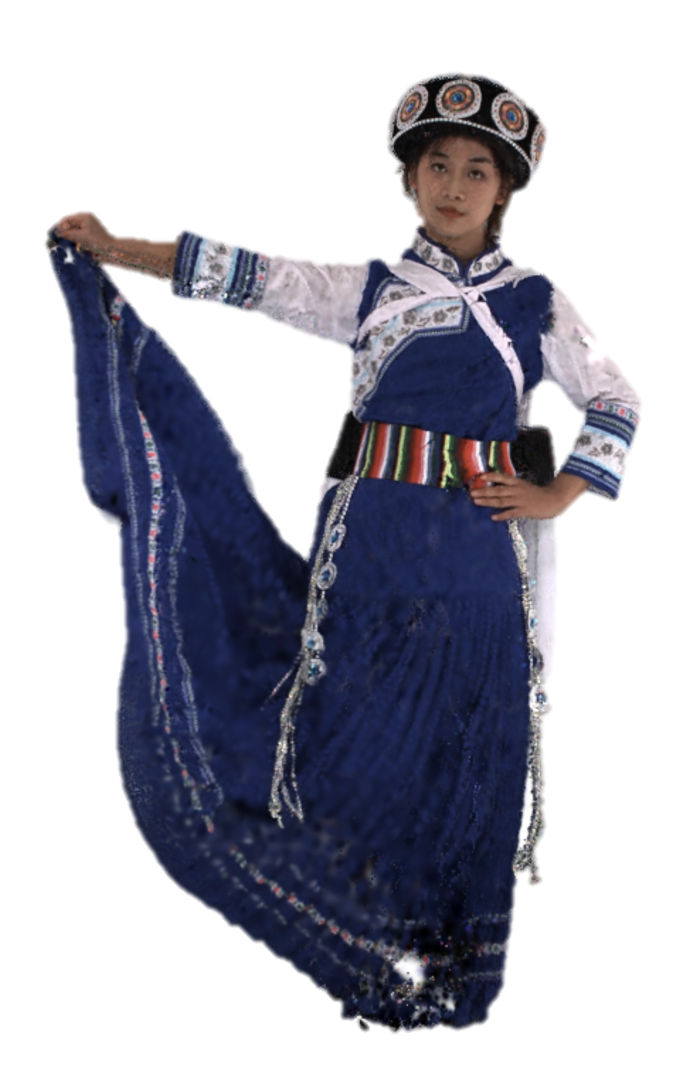}
        \caption{\hspace{1.1em}Ours} 
    \end{subfigure}
    \hfill
    \begin{subfigure}[h]{0.155\textwidth}
        \centering
        \includegraphics[width=0.8\textwidth]{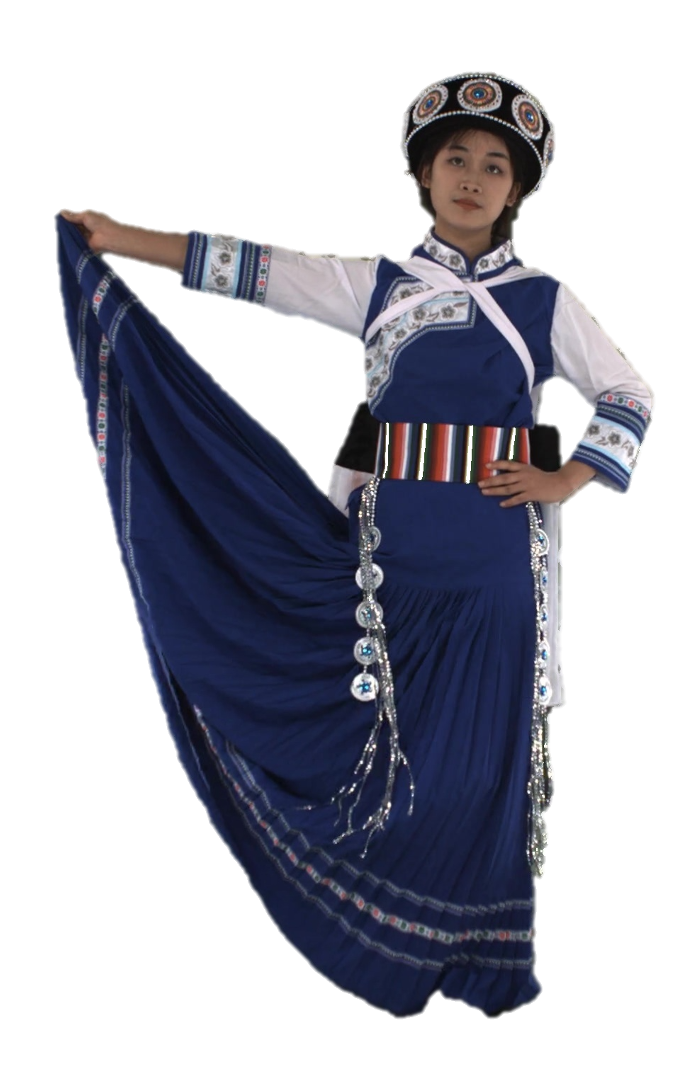}
        \caption{Ground Truth} 
    \end{subfigure}
    \par
    \vspace{10pt}
    \begin{subfigure}[h]{\linewidth}
        \scalebox{0.7}{
        \begin{tabular}{c|ccccc}
            \toprule
            Method & PSNR $\uparrow$ & SSIM $\uparrow$ & LPIPS $\downarrow$ & HF-SSIM $\uparrow$ & HF-PSNR $\uparrow$ \\
            \midrule
            AG\cite{li2024animatable} & 18.28 & \cellcolor{orange!25} 0.911 & \cellcolor{orange!25} 0.171 & 0.845 & 12.19 \\
            \textbf{PBDyG(Ours)} & \cellcolor{orange!25} 19.07 & 0.901 & 0.180 & \cellcolor{orange!25} 0.853 & \cellcolor{orange!25} 12.27\\
            \bottomrule
        \end{tabular}
        }
    \end{subfigure}
    \caption{Reconstruction comparison for Subject 0206. Our results have the same tendency (high qualitative accuracy, low quantitive results in SSIM and LPIPS).}
    \label{fig:0206-hf-ssim}
\end{figure}

\section{Experimental Results}

\begin{figure*}[t]
    \centering
    \includegraphics[width=\linewidth]{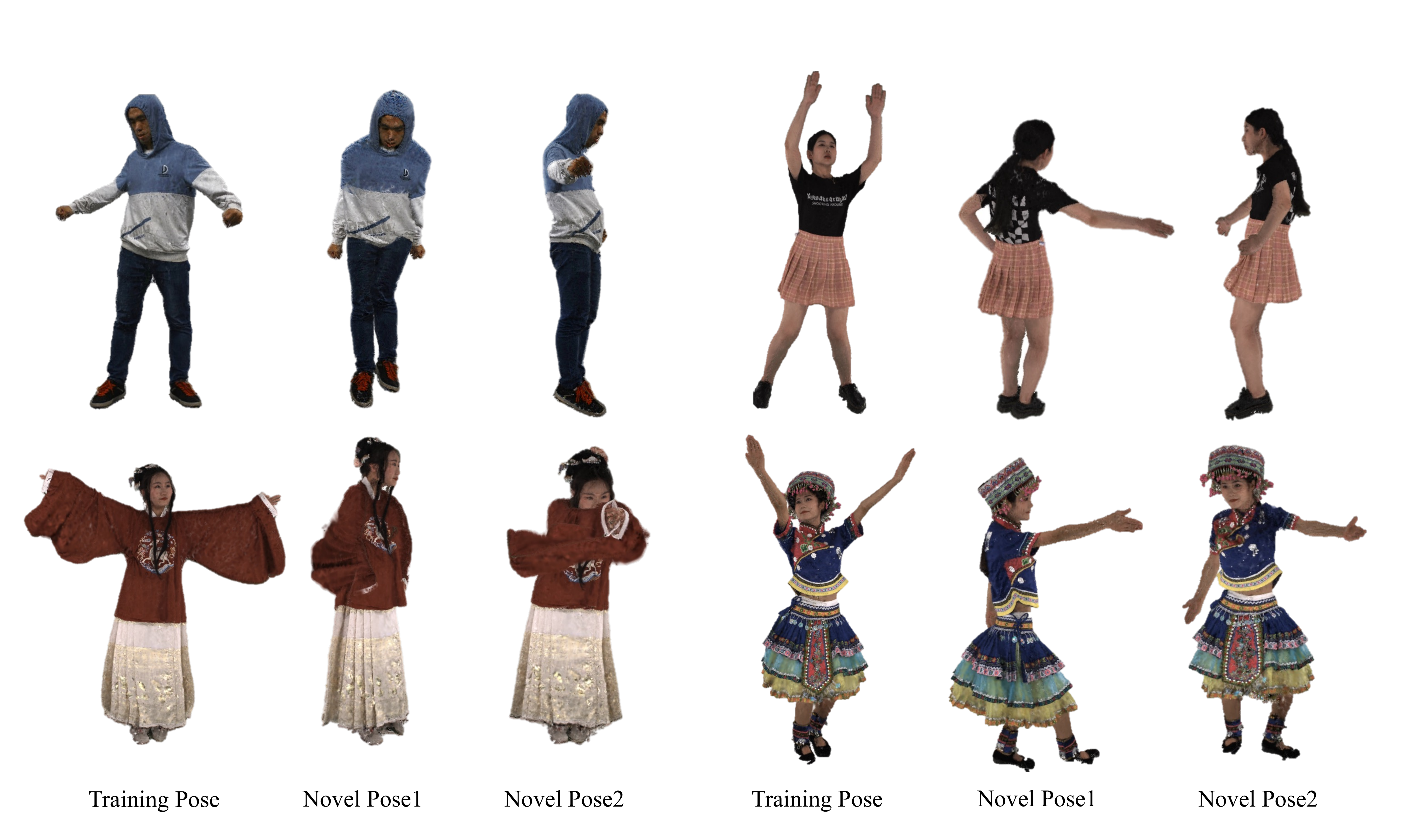}
    \caption{Example avatars reconstructed by our method. For each subject, the result on the left is in the training pose. The remaining columns are the avatars reanimated to two novel poses per example.}
    \label{fig:results}
\end{figure*}

We evaluate the effectiveness of our method on the DNA-rendering dataset which contains more challenging garments than traditional datasets.

\noindent\textbf{Quantitative Metrics.}
PSNR, SSIM, and LPIPS are commonly used evaluation metrics in many avatar restoration methods, but their scores does not always align with perceptual quality \cite{ZhangCVPR18}. To better measure the restoration of high-frequency information such as clothing patterns and wrinkles, we propose two new evaluation metric: High Frequency SSIM (HF-SSIM) and High Frequency PSNR (HF-PSNR).

In HF-SSIM, a high-pass filter is applied to both the Ground Truth and restored images to extract the high-frequency components, after which the SSIM metric is computed. We implemented the high-pass filter by removing the low-frequency components obtained via a Gaussian filter from the original image. In HF-SSIM, SSIM is calculated on the high-frequency components of the GT and restored images, making it more sensitive to whether the clothing details are accurately represented, which significantly influences the evaluation score. Additionally, since edges are emphasized, HF-SSIM applies a stronger penalty to ambiguous geometry. Through our experiments, we argue that existing metrics are insufficient for evaluating avatar restoration and that HF-SSIM provides a metric that is more aligned with human perception.

In HF-PSNR, the high-frequency components are extracted in the same way as in HF-SSIM, and then normal PSNR is calculated.

\noindent\textbf{Comparison with SOTA.}
We quantitatively compare our method with Animatable Gaussians \cite{li2024animatable} using subject0166, 0206, 0007 of DNA-Rendering dataset. We used the last 30 frames of the video as test data for each subject. During training, we select a frame for initialization, and the training data consists of the selected frame and the following 30 frames. Animatable Gaussians is also trained using the same training data. Our method depends not only on posture but also on the motion. We used the position and velocity of the last frame of the training data, obtained by PBD forward simulation, as the initial values during testing. Through our experiments, we found that the training of Animatable Gaussians becomes extremely unstable for certain highly challenging subjects. In cases where the loss becomes NaN or Inf during training, we used the network parameters just before this as the optimized result.

\cref{tab:comparison_results} shows the quantitative comparison results. At first glance, these results indicate that PBDyG has lower accuracy than the SOTA Animatable Gaussians. Close examination of the qualitative results together with these results prove otherwise and clearly show that our new metrics HF-SSIM and HF-PSNR are more appropriate to measure the quality of dynamic avatar reconstruction. 

\cref{fig:0166-hf-ssim} shows an example of restoring subject0166 from the DNA rendering dataset using Animatable Gaussians \cite{li2024animatable} and our method. We also present the results of calculating PSNR, SSIM, LPIPS, and our newly introduced HF-SSIM and HF-PSNR for these images. Qualitatively, Animatable Gaussians fails to reproduce the high-frequency components of clothing, such as the fine details and wrinkles. On the other hand, our method successfully restores the patterns on the clothing, even though PSNR and SSIM would indicate better results for Animatable Gaussians.

Next, in \cref{fig:0206-hf-ssim}, we present the results for subject0206. For this subject, Animatable Gaussians is able to restore the clothing details, but it fails to reconstruct the areas with significant motion-dependent deformation. Our method provides a more accurate geometry for these highly deformable areas, yet in terms of SSIM, and LPIPS, Animatable Gaussians still outperforms our approach.

\noindent\textbf{Qualitative Results.}
\cref{fig:results} shows results on other people. These results clearly show that our method can create accurate photorealistic human avatars with highly deformable clothes. The avatars reconstructed by our method can be driven by the body movements given as SMPL pose parameter sequence. Please see supplementary material for more results and animations.

\noindent\textbf{SMPL Refinement.}

\begin{figure}
\begin{center}
    \begin{subfigure}[h]{0.15\textwidth}
        \includegraphics[width=\textwidth]{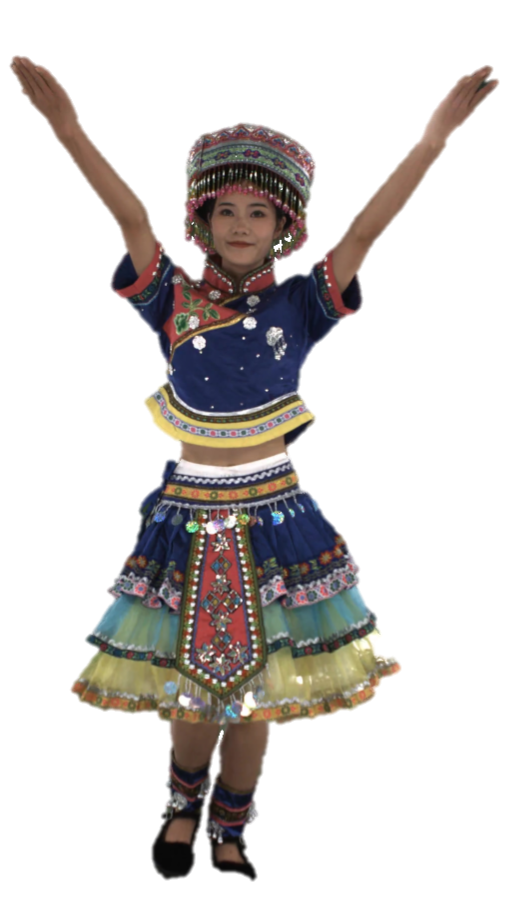}
        \caption{Reference View}  
    \end{subfigure}
    \begin{subfigure}[h]{0.15\textwidth}
        \includegraphics[width=\textwidth]{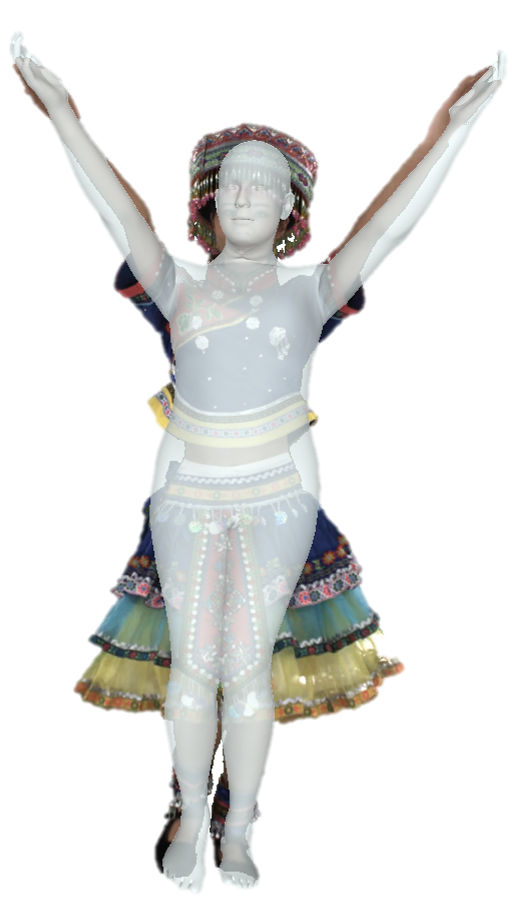}
        \caption{Original Annotation}  
    \end{subfigure}
    \begin{subfigure}[h]{0.15\textwidth}
        \includegraphics[width=\textwidth]{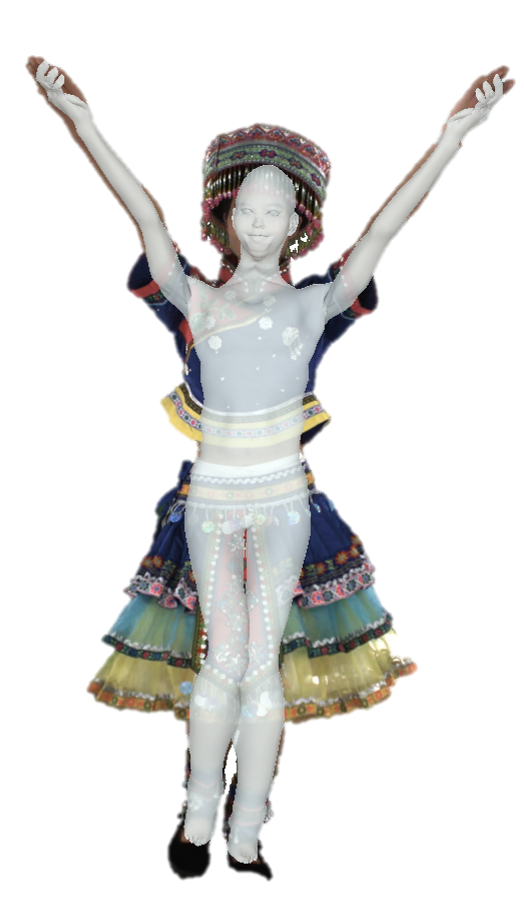}
        \caption{Ours}  
    \end{subfigure}
    \caption{SMPL refinement results. Our method references the 3D Gaussians to better fit the SMPL. The results clearly show the improvements.} 
    \label{fig:smpl_refine}
    \end{center}
    \vspace{-6mm}
\end{figure}

Our method leverages the 3D Gaussians to refine the SMPL fits. \cref{fig:smpl_refine} shows the results of this refinement. Our method is able to refine the original SMPL annotations of the dataset much more accurately. We believe this in itself would be a useful tool for many downstream applications of SMPL.

\begin{figure}
\begin{center}
    \begin{subfigure}[h]{0.15\textwidth}
        \includegraphics[width=\textwidth]{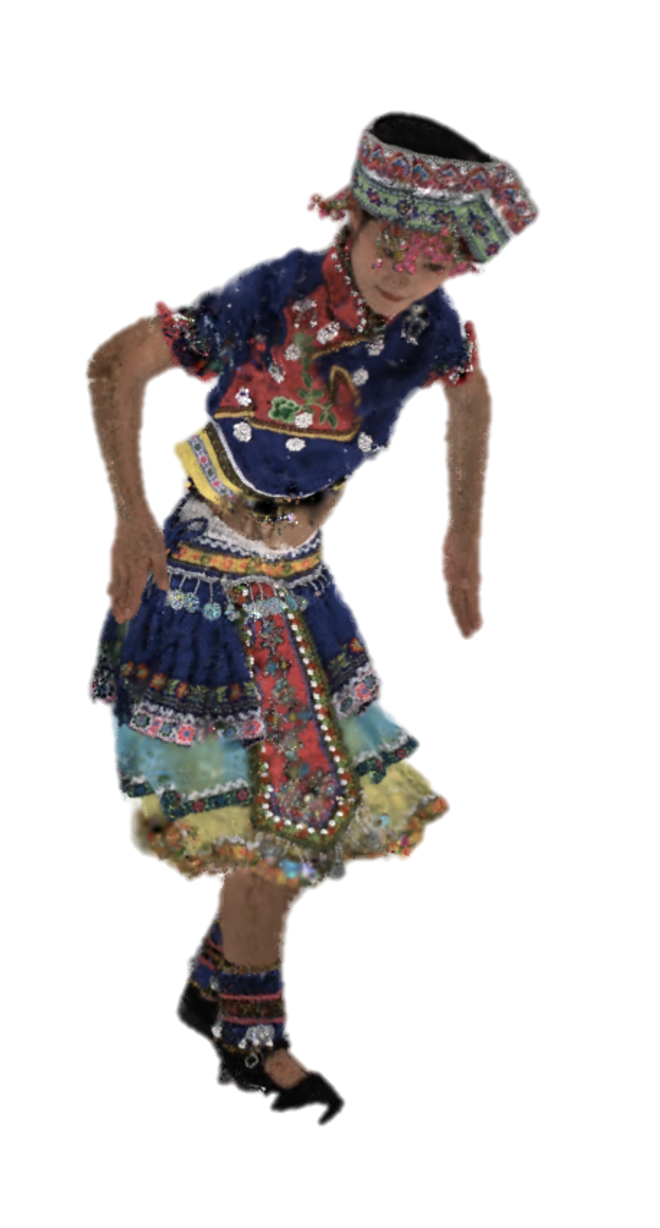}
        \caption{w/o AirMesh}  
    \end{subfigure}
    \begin{subfigure}[h]{0.15\textwidth}
        \includegraphics[width=\textwidth]{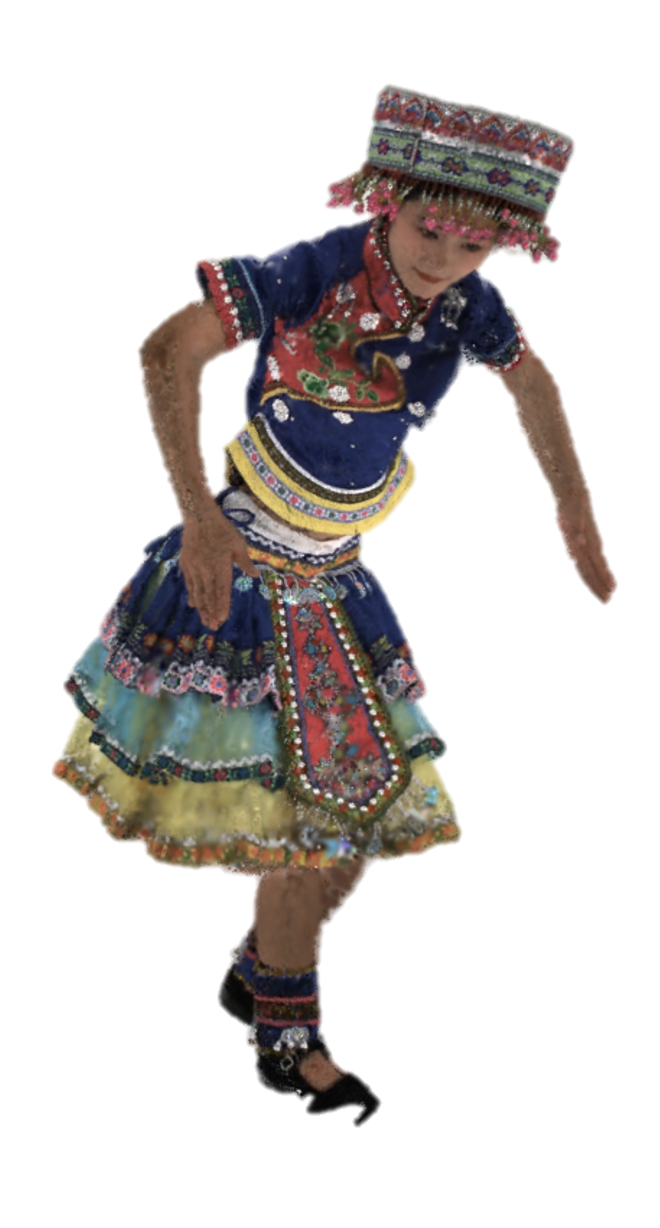}
        \caption{w/ AirMesh}  
    \end{subfigure}
    \begin{subfigure}[h]{0.15\textwidth}
        \includegraphics[width=\textwidth]{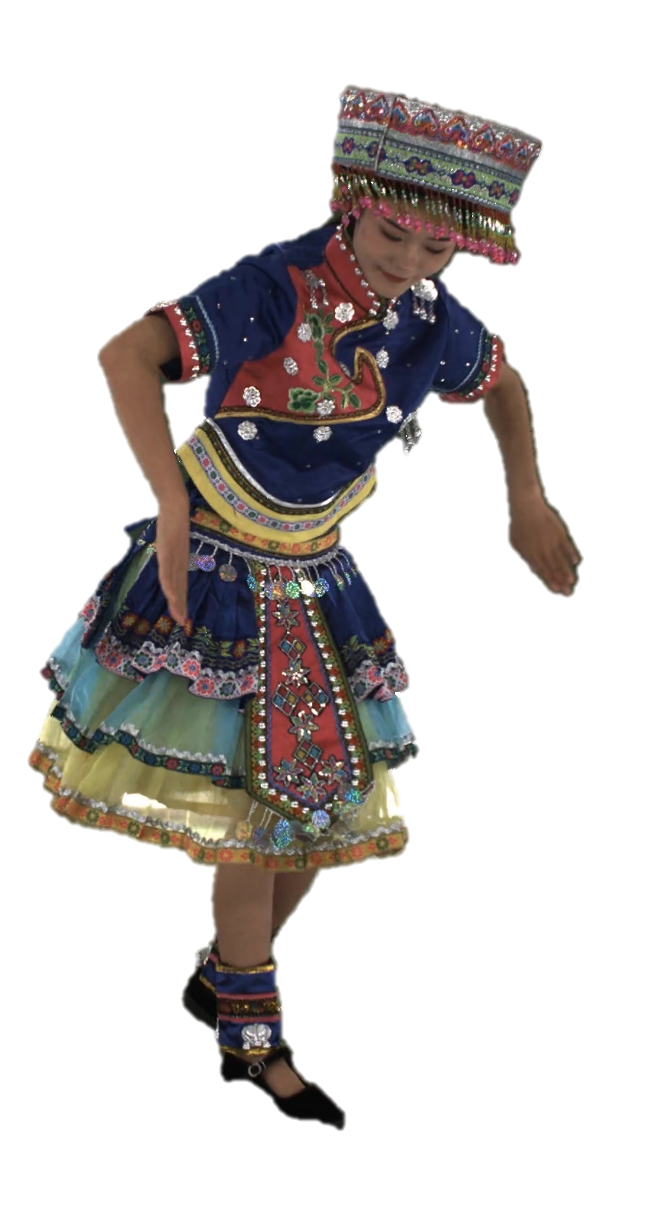}
        \caption{Ground Truth}  
    \end{subfigure}
    \caption{Ablation Study of AirMesh constraint in training pose. Without the AirMesh constraint, the shape deteriorates during the learning process.}
    \label{fig:airmesh-ablation}
    \end{center}
\end{figure}

\noindent\textbf{Ablation Studies.}
We empirically found that using distance constraints alone causes the set of Gaussians to collapse during the optimization of PBD parameters. To maintain the flexibility of the clothing while preventing the collapse of appearance and geometry, we introduce the AirMesh constraint \cite{muller2015airmesh} in addition to the distance constraints as a part of the PBD optimization. \cref{fig:airmesh-ablation} shows the results of ablating this AirMesh constraint. With the AirMesh constraint, we are able to optimize the PBD parameters while preserving both appearance and geometry during the training process.

\begin{figure}
\begin{center}
    \begin{subfigure}[h]{0.15\textwidth}
        \includegraphics[width=\textwidth]{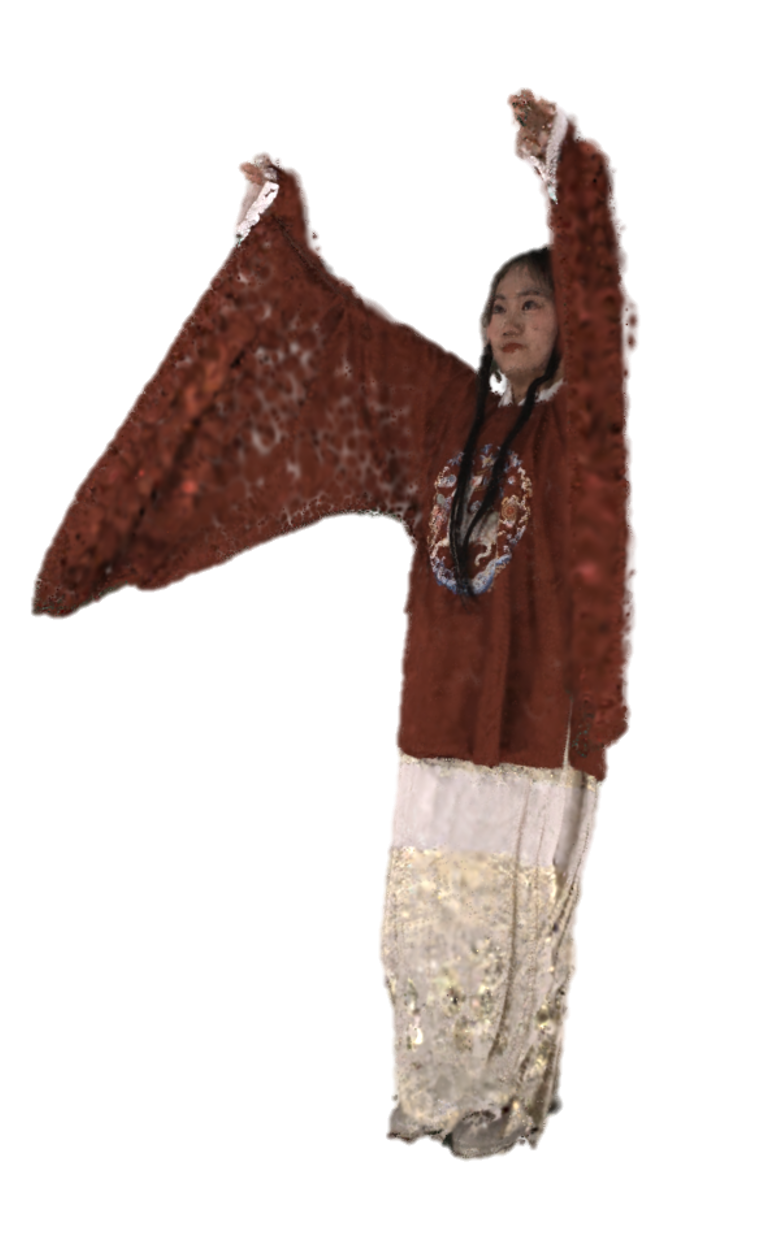}
        \caption{substep = 1}  
    \end{subfigure}
    \begin{subfigure}[h]{0.15\textwidth}
        \includegraphics[width=\textwidth]{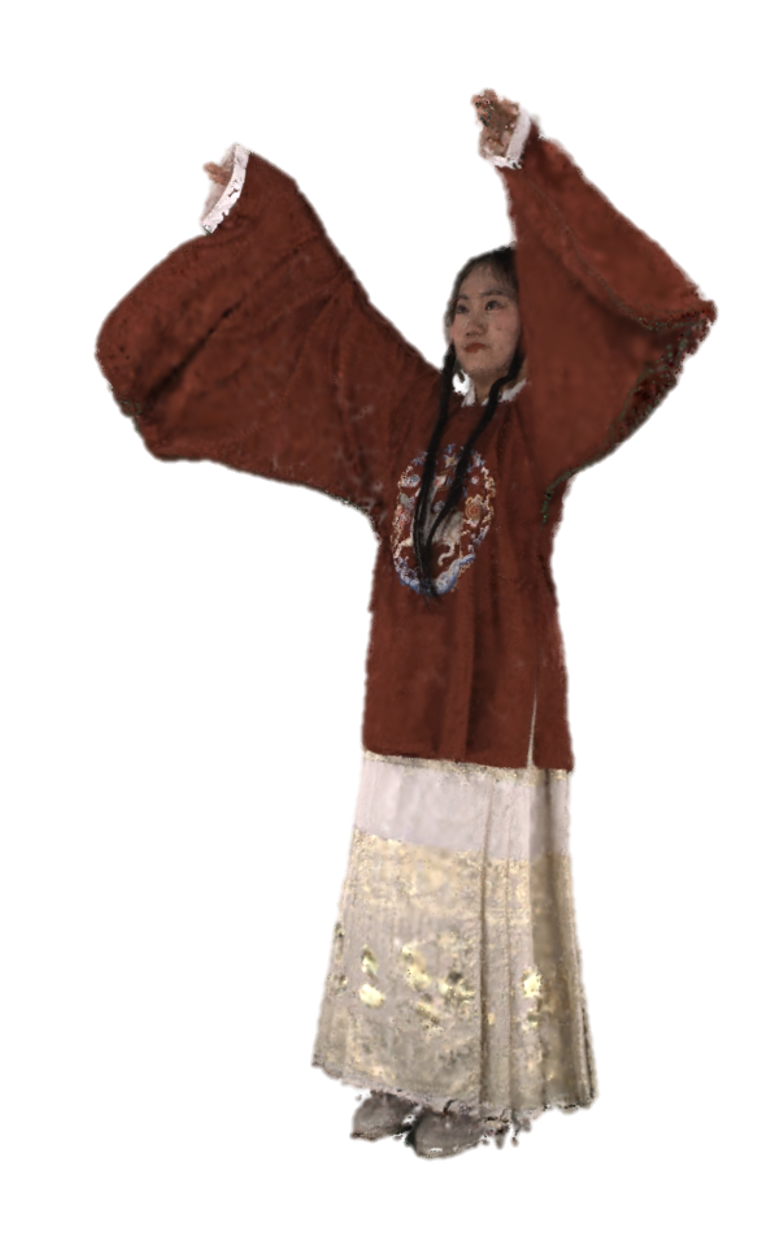}
        \caption{substep = 10}  
    \end{subfigure}
    \begin{subfigure}[h]{0.15\textwidth}
        \includegraphics[width=\textwidth]{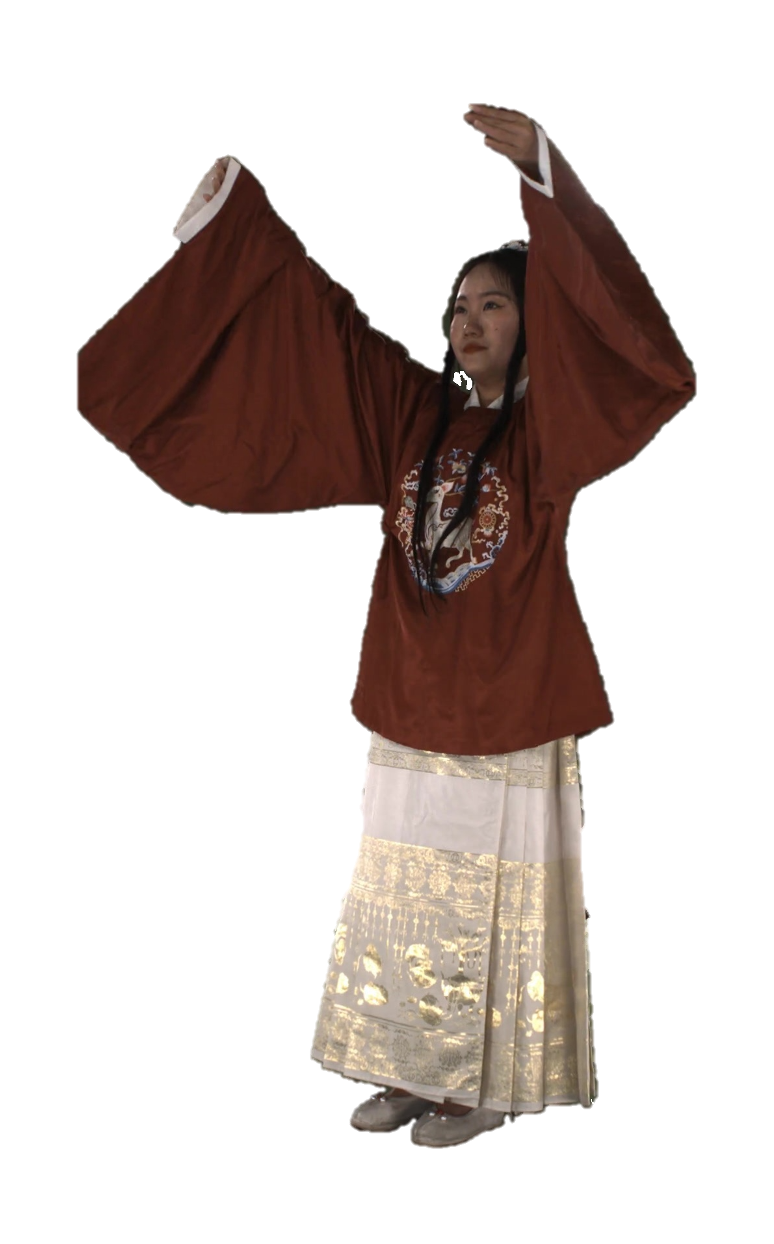}
        \caption{Ground Truth}  
    \end{subfigure}
    \caption{Ablation study of substep strategy. ``substep = k" means dividing the frame interval of the video into k substeps for simulation. By reducing the time span of the simulation, we can prevent the sagging caused by gravity.}
    \label{fig:substep-ablation}
    \end{center}
\end{figure}

When the time step is large, we are unable to correct the downward displacement caused by gravity, leading to sagging clothes. To achieve more accurate simulation, we divide each frame into 10 substeps and perform PBD at each substep. \cref{fig:substep-ablation} shows the results of ablating this the substep strategy. The results clearly show the importance of it.

\section{Conclusion}
We proposed PBDyG, a novel method for learning an animatable human avatar reconstruction from multiview videos. PBDyG models the human as 3D Gaussians whose movement-dependent motion can be simulated with position based dynamics driven by the underlying body modeled with SMPL. We showed that the physical parameters of the body and clothing can be learned from the video, which enables realistic animation of loose-clothed people with arbitrary movements. We believe PBDyG provides a sound integration of physically-based simulation in avatar recovery which may provide a convinient foundation for higher-level tasks in human behavior modeling and synthesis. 

\section*{Acknowledgements}
This work was in part supported by
JSPS 
20H05951 and 
21H04893, 
JST JPMJCR20G7 
and JPMJAP2305, 
and RIKEN GRP to KN, and 
the NSF Mathematical Sciences Postdoctoral Fellowship and the UC President's Postdoctoral Fellowship to JW.

\appendix

\section{Method Details}

\subsection{Delaunay Triangulation}

From a typical multiview image set of a person, 3D Gaussian Splatting\cite{kerbl20233d} reconstructs the person with roughly $10^4$ to $10^6$ Gaussians. The centers of these Gaussians form an unstructured 3D point cloud which cannot be directly simulated with Position Based Dynamics (PBD) \cite{muller2007position, macklin2016xpbd} due to the missing associations between the points. To apply PBD, the proposed method introduces connectivity between these Gaussians with Delaunay triangulation and tetrahedral subdivision.

Given a set of $n$ points $$ \mathcal{P} = \{ \mathbf{p}_1, \mathbf{p}_2, \dots, \mathbf{p}_n \} $$ of Gaussian centers, we first sample a subset $\mathcal{P}^{'}$ of 10,000 points.   With Delaunay triangulation applied to $\mathcal{P}'$, we obtain a set of  $m$ tetrahedra $\mathcal{T} = \{ T_1, T_2, \dots, T_m \}$ and a set of $l$ edges $\mathcal{E} = \{ E_1, E_2, \dots, E_l \}$, where each $ E_i $ represents an edge connecting two points in the triangulation.

This simple application of Delaunay triangulation generates a convex hull with many redundant tetrahedra. To remove unnecessary tetrahedra while maintaining the alignment of them with the shape of the person, we only retain those tetrahedra that are supported by their neighbors in their vicinity. For this, we first introduce the notion of local neighborhood with $k$-nearest neighbors, where $k=30$ in all our experiments. For each tetrahedron $T_i = \{\mathbf{v}_{i,1}, \mathbf{v}_{i,2}, \mathbf{v}_{i,3}, \mathbf{v}_{i,4}\}$, we first select one vertex, say $\mathbf{v}_{i,1}$, as the reference. We then compute the set of $k$-nearest neighbors of $\mathbf{v}_{i,1}$ from the point cloud $\mathcal{P}'$, denoted as $\mathcal{N}(\mathbf{v}_{i,1}, k)$.

Next, we check whether the remaining vertices $\mathbf{v}_{i,2}, \mathbf{v}_{i,3}, \mathbf{v}_{i,4}$ of tetrahedron $T_i$ are all contained within this neighborhood $\mathcal{N}(\mathbf{v}_{i,1}, k)$
\begin{equation}
\{ \mathbf{v}_{i,2}, \mathbf{v}_{i,3}, \mathbf{v}_{i,4} \} \subseteq \mathcal{N}(\mathbf{v}_{i,1}, 30)\,.
\end{equation}
If this condition is met, we include the tetrahedron $T_i$ in the valid tetrahedron set $\mathcal{T}_{\text{valid}}$. The set of valid tetrahedra $\mathcal{T}_{\text{valid}}$ becomes 
\begin{equation}
\mathcal{T}_{\text{valid}} = \{ T_i \in \mathcal{T} \mid \{ \mathbf{v}_{i,2}, \mathbf{v}_{i,3}, \mathbf{v}_{i,4} \} \subseteq \mathcal{N}(\mathbf{v}_{i,1}, 30) \}\,.
\end{equation}

By applying this criterion, we ensure that only those tetrahedra whose vertices are sufficiently close to each other, according to the defined neighborhood, are retained. The final set of tetrahedra $\mathcal{T}_{\text{valid}}$ aligns with the avatar shape without unnecessary redundancy.

\subsection{Interpolation}

The initial number of Gaussians recovered by 3D Gaussian Splatting applied to the input multiview images is prohibitively large. Direct use of all of them is simply impossible due to limitations in memory consumption and computational time. Instead, we perform PBD on a subset of 10,000 Gaussians which we obtain by sampling from the original set of Gaussians $\mathcal{P}$. The positions of the remaining Gaussians are interpolated from the result of PBD on this subset. We initially attempted to assign each Gaussian to a tetrahedron and perform barycentric interpolation. This, however, causes significant artifacts as some Gaussians do not belong to any tetrahedron and have negative barycentric coordinates. 

Instead, we introduce a custom interpolation algorithm to achieve high accuracy and to avoid unwanted artifacts. Let us denote the set of Gaussians not sampled as $\bar{\mathcal{P}} = \mathcal{P} - \mathcal{P}'$, where $\mathcal{P}$ is the original set of Gaussians and $\mathcal{P}'$ is the sampled subset used for the simulation. For each Gaussian in $\bar{\mathcal{P}}$, we apply Bounding Volume Hierarchy (BVH) to find the corresponding tetrahedron in the mesh. That is, we assign a tetrahedron to each in $\bar{\mathcal{P}}$, if the Gaussian lies within a tetrahedron $T_i \in \mathcal{T}_{\text{valid}}$, and its position is determined by barycentric interpolation within the tetrahedron $T_i$. If a Gaussian does not lie inside any tetrahedron, we compute the closest face of the closest tetrahedron. Let $F$ denote the closest face (a triangle) to the Gaussian, and let $d$ be the perpendicular distance from the Gaussian to the face. The barycentric coordinates $\mathbf{b} = (b_1, b_2, b_3)$ of the projected point on the face $F$ is assigned to this Gaussian as its relative coordinates.

The position of the Gaussian is then determined by applying barycentric interpolation on the deformed face $F$ using the barycentric coordinates $\mathbf{b}$, and then by translating along the normal direction of the face by the distance $d$. The position $\mathbf{p}_{\text{new}}$ of the Gaussian becomes
\begin{equation}
    \mathbf{p}_{\text{new}} = \sum_{k=1}^{3} b_k \mathbf{v}_k + d \mathbf{n}\,,
\end{equation}
where $\mathbf{v}_{k} (k = 1, 2, 3)$ are the vertices of the face $F$, $\mathbf{n}$ is the unit normal vector of the face $F$, $b_k$ is the barycentric coordinates of the projection of the Gaussian onto the face $F$, and $d$ is the perpendicular distance from the Gaussian to the face.

\subsection{Gaussian's Anisotropy During Deformation}

The Gaussians in 3D Gaussian Splatting are anisotropic in their orientations and shapes. When the Gaussians undergo deformation, it is crucial to appropriately preserve this anisotropy through the transformations. Since the set of Gaussians is sufficiently dense, we assume that, locally, the deformation of the Gaussians is rigid. For each Gaussian $\mathbf{p}_i \in \mathcal{P}$ in the initial frame, we first compute the three nearest neighbors $\mathbf{p}_{i1}, \mathbf{p}_{i2}, \mathbf{p}_{i3}$. The relative vectors $\mathbf{r}_{ij}$ between $\mathbf{p}_i$ and its neighbors $\mathbf{p}_{ij} (j = 1, 2, 3)$ are defined as 
\begin{equation}
\mathbf{r}_{ij} = \mathbf{p}_i - \mathbf{p}_{ij}, \quad \text{for} \quad j = 1, 2, 3 \,.
\end{equation}

We run the PBD iterations and at time $t$ after the tetrahedra are deformed, we compute the relative vectors $\mathbf{r}_{ij}(t)$ for the same Gaussian $\mathbf{p}_i$ and its neighbors. The matrix of these vectors at time $t$, $\mathbf{R}(t)$, becomes
\begin{equation}
\mathbf{R}(t) = \begin{bmatrix} \mathbf{r}_{i1}(t) & \mathbf{r}_{i2}(t) & \mathbf{r}_{i3}(t) \end{bmatrix}\,.
\end{equation}
The rotation matrix $\mathbf{X}$ that transforms the relative positions of the neighbors at time $t$ back to the initial configuration is 
\begin{equation}
\mathbf{X} = \mathbf{R}(t) \mathbf{R}^{-1}\,.
\end{equation}

To update the rotation of the Gaussian $\mathbf{p}_i$ at time $t$, we apply the rotation matrix $\mathbf{X}$ to the initial orientation $\mathbf{q}_i$ of the Gaussian. This yields the updated rotation $\mathbf{q}_i(t)$ at time $t$
\begin{equation}
\mathbf{q}_i(t) = \mathbf{X} \mathbf{q}_i\,.
\end{equation}

This computation allows us to track the rotation of each Gaussian, maintaining the correct orientation as the point cloud undergoes deformation. With this, we ensure that the anisotropic properties of the Gaussians are properly adjusted in response to the changes in their local environment.


{
    \small
    \bibliographystyle{ieeenat_fullname}
    \bibliography{main}
}

\end{document}